\documentclass[journal]{IEEEtran}

\usepackage{graphicx}
\usepackage{amsmath}
\usepackage{amsfonts}
\usepackage{mathtools}
\usepackage{amsmath}
\usepackage{diagbox}
\usepackage{multirow}
\usepackage{float}
\usepackage{color, soul}
\usepackage{icomma}
\usepackage[range-phrase=--]{siunitx}
\usepackage{algorithm, algpseudocode}
\usepackage{tabularx}
\usepackage{booktabs}

\usepackage{xargs}    
\usepackage[pdftex,dvipsnames]{xcolor} 
\usepackage[colorinlistoftodos,prependcaption,textsize=tiny]{todonotes}
\usepackage{multicol}

\newcommandx{\improvement}[2][1=]{\todo[linecolor=Plum,backgroundcolor=Plum!25,bordercolor=Plum,#1]{#2}}

\usepackage{glossaries}
\newacronym{MLC}{MLC}{multi-label scene classification}
\newacronym{SLC}{SLC}{single-label classification}
\newacronym{RS}{RS}{remote sensing}
\newacronym{CV}{CV}{computer vision}
\newacronym{LP}{LP}{label propagation}
\newacronym{xAI}{xAI}{explainable artificial intelligence}
\newacronym{GAN}{GAN}{generative adversarial network}
\newacronym{SAR}{SAR}{synthetic aperture radar}
\newacronym{CAMs}{CAMs}{class activation maps}

\makeatletter
\newcount\SOUL@minus 
\makeatother

\usepackage{etoolbox}
\usepackage{overpic}

\def\subcaptionbelow{subcaptionbelow}

\def\subcaptionstyle{subcaptionbelow}

\usepackage{cite}
\usepackage{hyperref}
\usepackage[
    singlelinecheck=false, 
    font=small,
]{caption}
\usepackage[slc=off, subrefformat=parens]{subcaption}

\ifx\subcaptionstyle\subcaptionbelow
\else
\fi

\usepackage[capitalise,noabbrev]{cleveref}

\begin{document}

\title{Annotation Cost-Efficient Active Learning for Deep Metric Learning Driven Remote Sensing Image Retrieval}

\author{Genc~Hoxha,~\IEEEmembership{~Member,~IEEE,}
       Gencer~Sumbul,~\IEEEmembership{~Member,~IEEE,}
       Julia Henkel,
       Lars Möllenbrok, ~\IEEEmembership{~Member,~IEEE,}
        and~Begüm~Demir,~\IEEEmembership{Senior~Member,~IEEE}
\thanks{Genc Hoxha, Julia Henkel, Lars Möllenbrok and Beg{\"u}m Demir are with the Faculty of Electrical Engineering and Computer Science, Technische Universit{\"a}t Berlin, 10623 Berlin, Germany and also with the Berlin Institute for the Foundations of Learning and Data (BIFOLD), 10623 Berlin, Germany (emails: genc.hoxha@tu-berlin.de, henkel@campus.tu-berlin.de lars.moellenbrok@tu-berlin.de, demir@tu-berlin.de).

Gencer Sumbul is with the Environmental Computational Science and Earth Observation Laboratory (ECEO), École Polytechnique Fédérale de Lausanne (EPFL), 1950 Sion, Switzerland (e-mail: \mbox{gencer.sumbul@epfl.ch}).} 

}

\markboth{Journal of \LaTeX\ Class Files,~Vol.~14, No.~8, August~2015}%
{Shell \MakeLowercase{\textit{et al.}}: Bare Demo of IEEEtran.cls for IEEE Journals}

\maketitle

\begin{abstract}
Deep metric learning (DML) has shown to be effective for content-based image retrieval (CBIR) in remote sensing (RS). Most of DML methods for CBIR rely on a high number of annotated images to accurately learn model parameters of deep neural networks (DNNs). However, gathering such data is time-consuming and costly. To address this, we propose an annotation cost-efficient active learning (ANNEAL) method tailored to DML-driven CBIR in RS. ANNEAL aims to create a small but informative training set made up of similar and dissimilar image pairs to be utilized for accurately learning a metric space. The informativeness of image pairs is evaluated by combining uncertainty and diversity criteria. To assess the uncertainty of image pairs, we introduce two algorithms: 1) metric-guided uncertainty estimation (MGUE); and 2) binary classifier guided uncertainty estimation (BCGUE). MGUE algorithm automatically estimates a threshold value that acts as a boundary between similar and dissimilar image pairs based on the distances in the metric space. The closer the similarity between image pairs is to the estimated threshold value the higher their uncertainty. BCGUE algorithm estimates the uncertainty of the image pairs based on the confidence of the classifier in assigning correct similarity labels. The diversity criterion is assessed through a clustering-based strategy. ANNEAL combines either MGUE or BCGUE algorithm with the clustering-based strategy to select the most informative image pairs, which are then labelled by expert annotators as similar or dissimilar. This way of annotating images significantly reduces the annotation cost compared to annotating images with land-use land-cover class labels. Experimental results on two RS benchmark datasets demonstrate the effectiveness of our method. The code of this work is publicly available at \url{https://git.tu-berlin.de/rsim/anneal_tgrs}.
\end{abstract}

\begin{IEEEkeywords}
Active learning, content-based image retrieval, deep metric learning, remote sensing.
\end{IEEEkeywords}

\IEEEpeerreviewmaketitle

\section{Introduction}\label{sec:introduction}
\IEEEPARstart{W}{ith} the rapid development of remote sensing (RS) technology, we have witnessed an unprecedented growth in the volume of RS image archives. Accordingly, one of the most important research topics in RS is the development of fast and accurate content based image retrieval (CBIR) methods. CBIR methods aim at searching and retrieving semantically similar images to a user-defined query image from massive archives \cite{Sumbul:2021, surveyCBIR2021,investig_dhatcu2019,CBIR_SURVEY_23}. This is usually achieved based on two main steps. The first step is devoted to the characterization of the complex content of a RS image scene with a set of discriminitave features, whereas the second step is devoted to the retrieval of the most similar images by evaluating the similarity between the query image and those of an archive in the feature space. In particular, the accurate characterization of images is crucial to reach an accurate search capability within huge data archives. In this context, deep metric learning (DML) has recently shown to be very effective for CBIR in RS. DML aims at learning a feature (i.e., metric) space, where similar images are mapped close to each other and dissimilar samples are mapped apart from each other. The amount and quality of the available training samples (i.e., images) are important to learn an accurate metric space. However, in operational scenarios gathering a sufficient number of labeled training images is not realistic due to the high cost and the related time consuming process of this task. To deal with this problem, active learning (AL) methods have been presented in the literature in the context of CBIR \cite{ferecatu2007interactive_AL_CBIR,demir2014novel_AL_CBIR}. AL aims to expand an initial and sub-optimal training set in an iterative manner by finding the most informative images from an archive that when annotated and included in the training set can significantly improve the retrieval performance\cite{demir2014novel_AL_CBIR}. The informativeness of the images is usually assessed based on two criteria: 1) uncertainty; and 2) diversity. 
Given a query image, the uncertainty criterion aims at finding images for which the confidence of a supervised algorithm in correctly classifying them as relevant or irrelevant to the given query image is low. The diversity criterion aims at finding diverse images to reduce the redundancy of the selected images as much as possible.
As an example, Feratcatu et al. \cite{ferecatu2007interactive_AL_CBIR} present an AL method in the context of binary support vector machines (SVMs) \cite{vapnik1999nature,SVM_Melg_Bruzz_04} that combines the two aforementioned criteria to select the most informative images based on two steps. In the first step, the most uncertain images are selected from the archive based on the  margin sampling (MS) strategy \cite{schohn2000less,ALtext_SVM}. In the second step, the most diverse images among the uncertain ones are defined by selecting the most distant samples in the feature space. In \cite{demir2014novel_AL_CBIR} a triple criteria AL method is proposed that includes a density criterion in addition to the uncertainty and the diversity criteria. The three criteria are jointly evaluated in two steps. In the first step, similarly as in\cite{ferecatu2007interactive_AL_CBIR}, the most uncertain images from the archive are selected based on the MS strategy. In  the second step the most diverse images among the uncertain ones are selected from the highest density regions in the feature space. This is achieved using kernel k-means clustering technique to cluster the uncertain images into different clusters in the feature space. Then, the most representative image from each cluster is selected as the one situated in the highest density region of the respective cluster.

\begin{figure}[!tbp]
  \begin{subfigure}[b]{0.48\textwidth}
    \includegraphics[width=\textwidth]{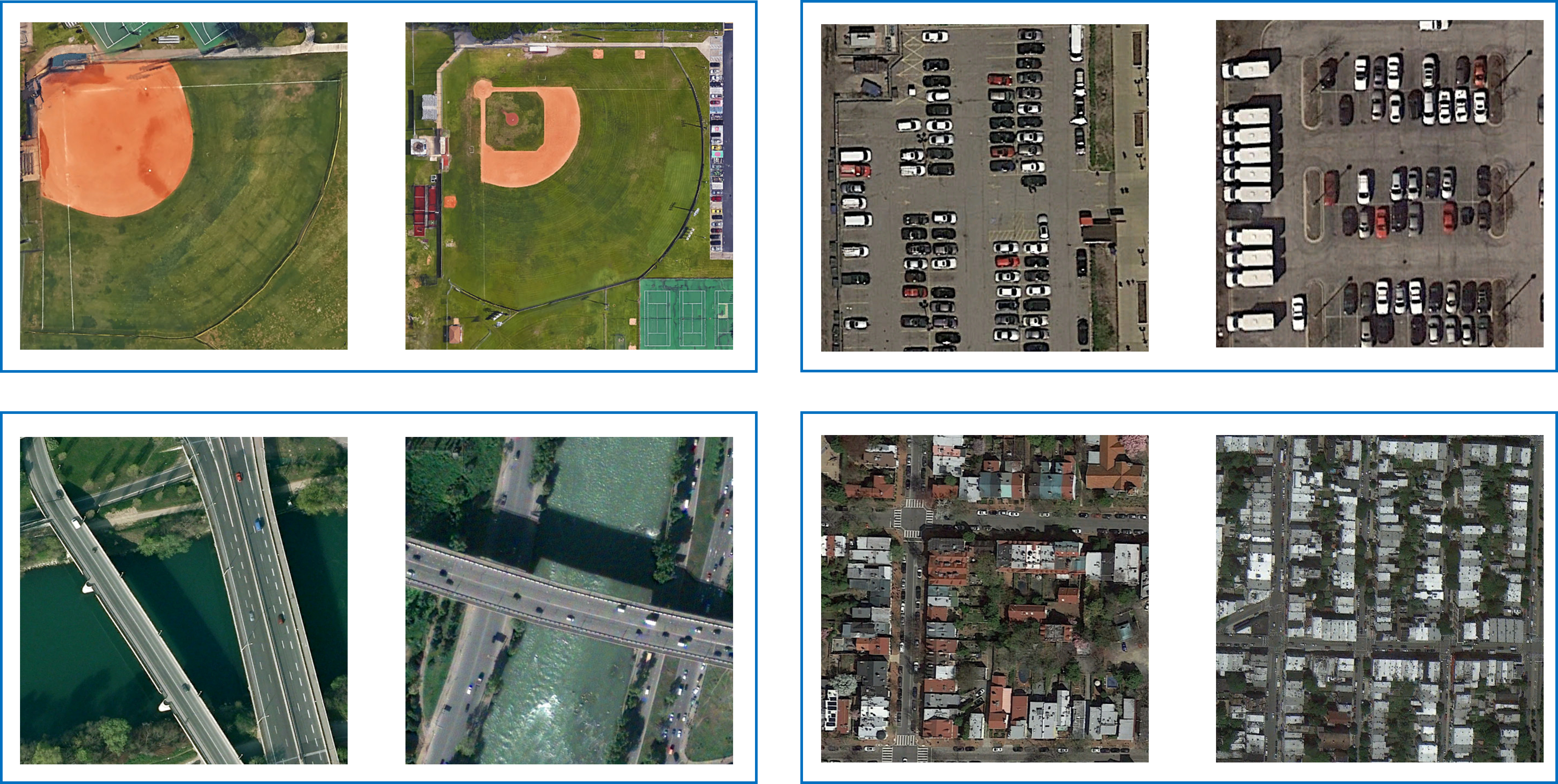}
    \caption{}
    \label{fig:similar}
  \end{subfigure}
  \hfill
  \begin{subfigure}[b]{0.48\textwidth}
    \includegraphics[width=\textwidth]{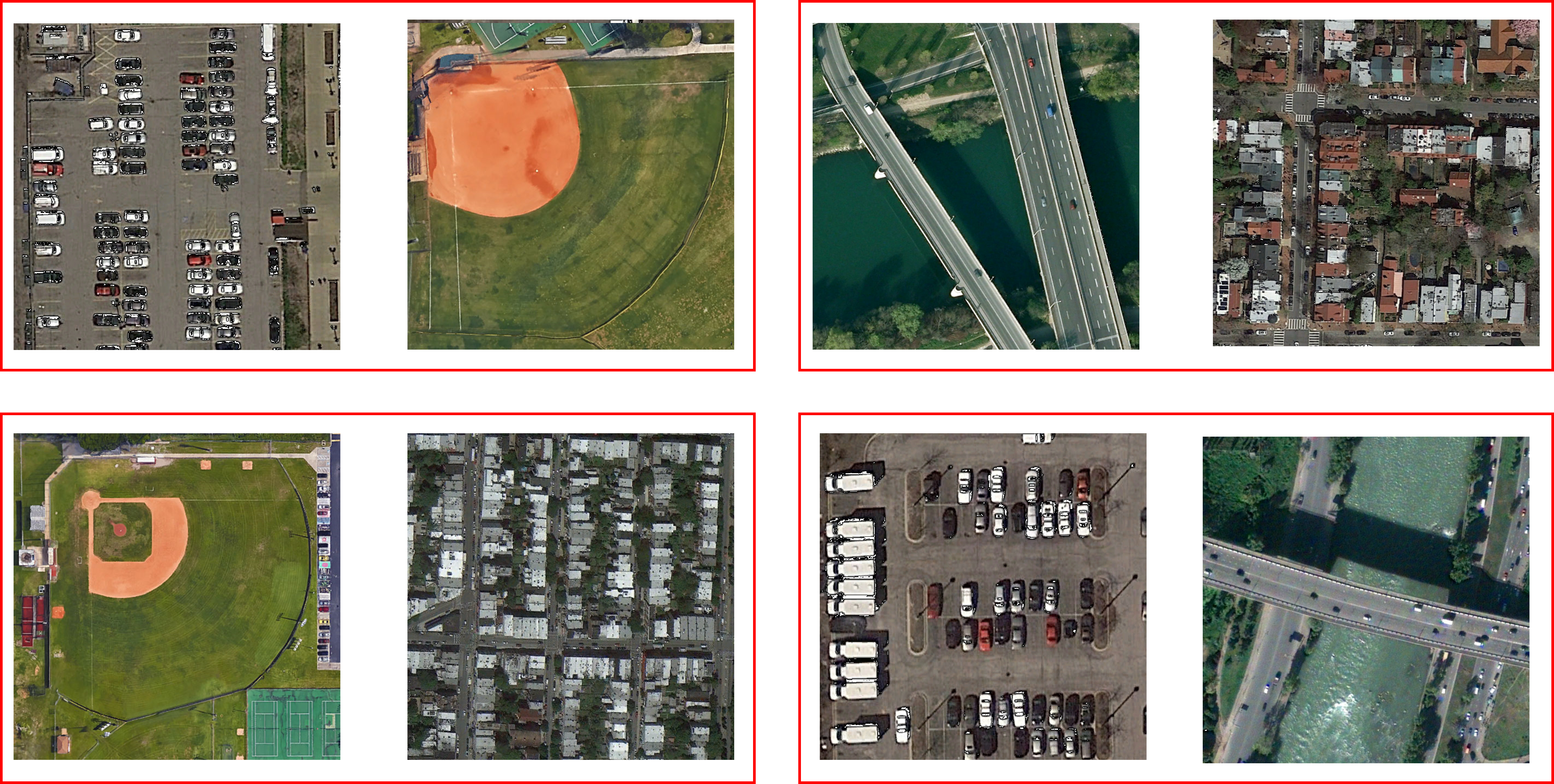}
    \caption{}
    \label{fig:dissimilar}
  \end{subfigure}
  \caption{Examples of image pairs from AID dataset \cite{xia2017aid}: a) similar pairs (in blue frame) and b) dissimilar pairs (in red frame).}
  \label{fig:AID}
\end{figure}
The aforementioned AL methods have shown their effectiveness in reducing the need of annotating a high number of images. However, they result in limited retrieval performance as they are defined in the context of traditional classifiers such as SVMs that rely on hand-crafted features. Moreover, the CBIR systems with these AL methods require a new AL process when a new query image is selected. This involves: i) asking the human expert to annotate the selected images as relevant or irrelevant with respect to the considered query image; ii) training a separate (ad-hoc) binary classifier using the annotated images with two class labels (i.e., relevant/irrelevant) with respect to the query image; and iii) repeating the process until the user is satisfied with the results or a predefined labeling budged is reached. This procedure is intractable in operational scenarios as the number of query images and their diversity can be high. As a result, AL methods that allow the creation of a training set with informative images independent from the selected query images are needed. To this end, AL methods proposed for classification problems with DNNs \cite{sener2018active, moellenbrokGRSL,moellenbrokIGARSS} could be used to simultaneously construct informative training sets and learn image representations. However, in this way image representations are encoded to discriminate land use land cover (LULC) classes rather than modeling semantic similarities. Furthermore, the collection of LULC class labels is complex and time-consuming, especially when dealing with large number of LULC class labels \cite{hu2020one_bit_supervision,zhang2022one_bit_supervision}.

To overcome these issues, in this paper we introduce an \textbf{ann}otation cost-\textbf{e}fficient \textbf{AL} method (\textbf{ANNEAL}) for DML driven CBIR in RS. ANNEAL aims to construct a small but informative training set made up of similar and dissimilar RS image pairs to be utilized for accurately and efficiently learning a deep metric space. To this end, it is defined based on a two-step procedure: 1) the selection of the most uncertain image pairs; and 2) the selection of the most diverse image pairs among the most uncertain ones. For the first step, we propose two different algorithms: 1) a metric-guided uncertainty estimation (MGUE); and 2) a binary classifier guided uncertainty estimation algorithm (BCGUE). The first algorithm employs uncertainty assessment of image pairs directly in a deep metric space as the first time in AL literature. This is achieved based on the automatic estimation of an adequate threshold value to distinguish between similar and dissimilar image pairs. The unlabeled image pairs that have a similarity value closest to the threshold value are the ones characterized by the highest degree of uncertainty. The estimation of the threshold value is based on the distance in the feature (i.e., metric) space between the images that compose the similar and dissimilar pairs in the current training set. In the second algorithm, the uncertainty of the image pairs is assessed based on the confidence of the classifier in assigning the correct similarity label that requires the incorporation of a binary classifier in the considered CBIR system. As the confidence of the classifier becomes lower, the uncertainty of the considered image pair becomes higher. In the second step, we select the most diverse image pairs among the most uncertain ones through a clustering based strategy. In detail, the $k$-means clustering algorithm is utilized to first cluster the most uncertain pairs into different clusters, and then we select one image pair per cluster. ANNEAL is defined as the combination of one of the proposed uncertainty with the clustering-based strategy. We call ANNEAL with MGUE algorithm as ANNEAL-MGUE, while that with BCGUE algorithm as ANNEAL-BCGUE. The selected most informative pairs either by ANNEAL-MGUE or ANNEAL-BCGUE are labeled by a human annotator as similar/dissimilar with respect to each other. Defining the similarity of two RS images based on their contents naturally aligns with the goal of CBIR. Fig. {\ref{fig:AID}} depicts some examples of similar and dissimilar RS image pairs. Based on the visual inspection of such pairs, a human expert is required to define their similarity. Annotating images as similar/dissimilar significantly reduces the annotation cost compared to annotating images with LULC class labels. Referring to the information theory, it requires only one bit of information to label a pair of images (i.e., one sample) as similar/dissimilar. In contrast, $\text{log}_2{C}$ bits of information are required to label an image with one of the predefined $C$ LULC class labels. Moreover, annotating images as similar/dissimilar allows to obtain further training pairs with zero cost due to transitive property of similarity \cite{roy2018exploiting_transitive}.

We would like to note that BCGUE has been briefly presented in \cite{henkel2023annotation} with limited experimental analysis. The contribution of this paper, which significantly extends our previous work presented in \mbox{\cite{henkel2023annotation}}, consists in: 1) the detailed description of ANNEAL-BCGUE with enriched experiments; 2) introducing ANNEAL-MGUE (which does not require the use of a binary classifier in the considered CBIR system); and 3) comparing theoretically and experimentally ANNEAL-BCGUE and ANNEAL-MGUE with an extended experimental analysis on two RS benchmark datasets. Experimental results show the effectiveness of ANNEAL in general. Among the proposed uncertainty evaluation algorithms, we observe that MGUE leads to more accurate uncertainty estimation of the image pairs directly on a metric space, while reducing the model complexity in terms of the number of parameters.

The remaining part of this paper is organized as follows. Section \ref{related_CBIR} presents the related work on CBIR in RS. Section \ref{methods} introduces the proposed method. Section \ref{dataset_setup} describes the considered RS image archives and the experimental setup, while the experimental results are presented in Section \ref{experimental_results}. Finally, in Section \ref{conclusion}, the conclusion of the work is drawn.
\section{Related Work on DML for CBIR in RS} \label{related_CBIR}

The development of DML methods, which learn a feature (i.e., metric) space in which similar images are close to each other and vice versa, has attracted attention in RS due to their capabilities to model image similarities for accurate CBIR \cite{Sumbul:2021}. As an example, Chaundhuri et al. \cite{chaudhuri2019siamese_graph_CNN_DML_PAIR} introduce a DML-based CBIR system for very high resolution (VHR) images that employs a Siamese graph CNN to learn image similarity from region adjacency graphs (RAGs) using contrastive loss function \cite{LeCun_loss_function_2006}. The authors create mini-batches of tuples of similar and dissimilar images based on the class labels (i.e., if the images share the same class label they are considered as similar and vice versa). Subsequently, RAG representations are fed into two graph CNNs with shared weights that are trained through contrastive loss to model image similarity \cite{LeCun_loss_function_2006}. A triplet CNN network with shared weights is proposed in \cite{triplet_enhancing2020}, where image triplets are used instead of tuples to learn image similarities. The triplets consist of: a) an anchor image; b) a positive image that is similar (i.e., share the same class label) to the anchor; and c) a negative image that is dissimilar (i.e., different class label) to the anchor. The triplet loss \cite{schroff2015facenet_triplet_loss} is used to train the network, aiming to minimize the distance in the feature space between the anchor and the positive pair, while maximizing that of the anchor and the negative pair. Imbriacco et al. \cite{triplet_imbriaco_20} present a CBIR system based on a multi-branch CNN to extract global and local features combining the triplet loss \cite{schroff2015facenet_triplet_loss} with the cross-entropy loss. A global optimal structured loss function that uses image triplets is proposed in \cite{liu2020global} to produce an embedding space in which the positive pairs are forced to stay in a compact form within a hyper-sphere while the negative ones far away from its boundaries. To this end, the authors in \cite{liu2020global} combine the global lifted structure loss \cite{hermans2017defense_global_lifted} and the cross-entropy loss to learn global discriminative features with the hardest positive and negative mining strategy \cite{wang2019multi_mining_strategy} that increases intra-class compactness and inter-class sparsity. Similarly, Fan et al. \cite{fan2020distribution} propose a distribution consistency loss function that uses multiple positive and negative images for each anchor (unlike the  triplet loss function) considering the equal distribution of the mined easy and hard samples within each class. The  distribution consistency loss function is a combination of sample balance loss and the ranking consistency loss \cite{distriub_consist_loss_trip}. 
The former assigns dynamic weights to the selected hard samples based on the ratio of hard and easy samples in a given class, while the latter ranks the negative samples according to their category distribution. Sumbul et al \cite{Triplet_sumub_22_DML} present a  triplet selection strategy to select informative and representative triplets based on relevancy, hardness and diversity of the selected images. DML has been also successfully applied in the context hashing-based CBIR systems \cite{hasing_method_hand,Hashing_roy_21_triplet,self_sup_hashing_22_pairs,sup_pairWise_contrastive_23}. As an example, Roy et al. \cite{Hashing_roy_21_triplet} present a DML based hashing method that combines the triplet loss with bit-balancing and push loss \cite{yang2017supervised_push_loss_bit_balance} to generate discriminative hash codes for CBIR.  Similarly, to obtain discriminative hash code representations, Li et al. \cite{hasing_method_hand} combine contrastive and bit-balance loss functions. For a comprehensive and an extensive survey on the recent advancement of hashing-based CBIR systems in RS we refer the reader to the recent review paper proposed by Zhou et al. \cite{CBIR_SURVEY_23}.
\section{Proposed Annotation Cost-Efficient Active Learning (ANNEAL) Method}\label{methods}
\begin{figure*}
    \centering
    \includegraphics[scale =0.6]{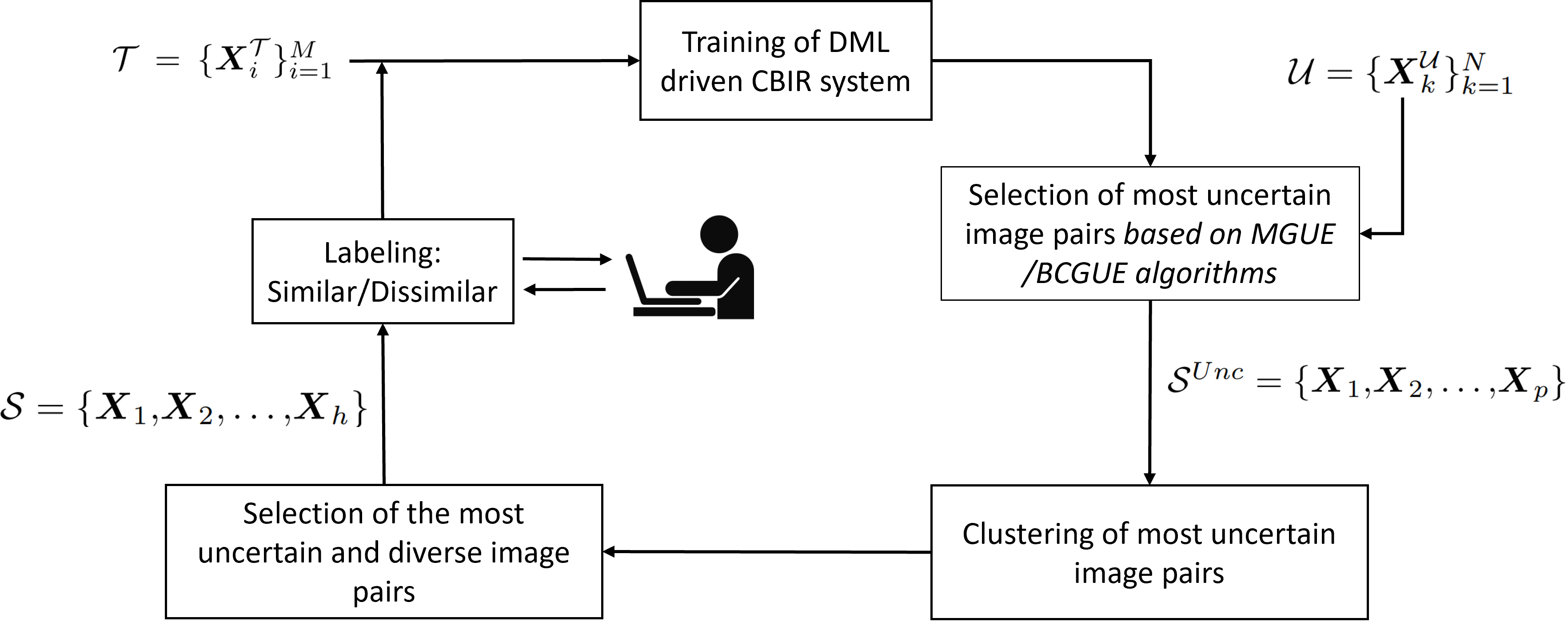}
    \caption{Block diagram of the proposed ANNEAL method.}
    \label{Fig:AL}
\end{figure*}

Let $\mathcal{I}$ be an RS image archive and $\mathcal{T} = \{\boldsymbol{X}_i^\mathcal{T}\}^M_{i=1}$ be an initial training set made up of $M$ RS image pairs, where $\boldsymbol{X}_i^\mathcal{T}$ is the i-th pair composed of images $\boldsymbol{I}^{i,\mathcal{T}}_1$ and $\boldsymbol{I}^{i,\mathcal{T}}_2$ from $\mathcal{I}$, i.e., $ \forall \ \boldsymbol{X}_i^\mathcal{T} = (\boldsymbol{I}^{i,\mathcal{T}}_1,\boldsymbol{I}^{i,\mathcal{T}}_2) \in \mathcal{T} \!\!\!\quad \exists \!\!\!\quad \boldsymbol{I}^{i,\mathcal{T}}_1, \boldsymbol{I}^{i,\mathcal{T}}_2 \in \mathcal{I}$. Each image pair $\boldsymbol{X}_i^\mathcal{T}$ is associated with a similarity label $\boldsymbol{y}_i^{\mathcal{T}}$, where $\boldsymbol{y}_i^{\mathcal{T}} = 1$ if $\boldsymbol{I}^{i,{\mathcal{T}}}_1$ and $\boldsymbol{I}^{i,\mathcal{T}}_2$ are similar to each other and $\boldsymbol{y}_i^{\mathcal{T}} = 0$ otherwise. Let $\boldsymbol{X}_k^{\mathcal{U}}$ be an unlabeled image pair and $\mathcal{U}=\{ \boldsymbol{X}_k^{\mathcal{U}}\}^N_{k=1} $ be a set of $N$ unlabeled pairs, where $N \gg M$. We assume that the number $|\mathcal{T}|$ of labeled training image pairs is not sufficient for accurately learning a deep metric space. To construct a small but informative training set of similar and dissimilar RS image pairs, we introduce an annotation cost-efficient AL (ANNEAL) method for DML driven CBIR in RS. To this end, ANNEAL iteratively selects a batch $\mathcal{S}=\{\boldsymbol{X}_1^{\mathcal{U}},\boldsymbol{X}_2^{\mathcal{U}},\ldots,\boldsymbol{X}_h^{\mathcal{U}}\}$ of $h$ most informative image pairs from $\mathcal{U}$ that are uncertain (i.e., ambiguous) and as much diverse as possible to each other. This is achieved based on a two-step procedure. In the first step, ANNEAL selects the set $\mathcal{S}^{Unc}=\{\boldsymbol{X}_1,\boldsymbol{X}_2,\ldots,\boldsymbol{X}_p\}$ of $p$ most uncertain image pairs with $p>h$. In the second step, the $h$ most diverse samples are selected from the most uncertain set $\mathcal{S}^{Unc}$. The selected most informative image pairs are annotated by human experts as similar or dissimiliar and added to the current training set. The iterative process is terminated based on CBIR performance or labeling budget. We would like to note that this way of annotating images reduces the annotation cost for a given sample (i.e., image pair) to only one bit of information ($log_2{2} = 1$) compared to associating each image with one of $C$ LULC class labels (which require $log_2{C}$ bits of information). Before explaining in detail the proposed ANNEAL method (which is illustrated in {\ref{Fig:AL}}), we first present the considered DML based CBIR system in the following.

\subsection{The Considered DML Driven CBIR System}
 The CBIR system considered in this work is based on DML. As in any CBIR system, one of the two fundamental steps is the image characterization with discriminative features. To this end, we consider two DNNs with shared weights (i.e., Siamese neural network [SNN]) that are trained using the contrastive loss \cite{LeCun_loss_function_2006} for learning a metric space. To this end, each image pair $\boldsymbol{X}_i^\mathcal{T} = (\boldsymbol{I}^{i,\mathcal{T}}_1,\boldsymbol{I}^{i,\mathcal{T}}_2)\in \mathcal{T}$ is given as input to the SNN to obtain corresponding image features $ \boldsymbol{F}_i^\mathcal{T} = (\boldsymbol{f}^{i,\mathcal{T}}_1,\boldsymbol{f}^{i,\mathcal{T}}_2)$, where $\boldsymbol{f}^{i,\mathcal{T}}_1$ and $\boldsymbol{f}^{i,\mathcal{T}}_2$ denote the image features of $\boldsymbol{I}^{i,\mathcal{T}}_1$ and $\boldsymbol{I}^{i,\mathcal{T}}_2$, respectively. Then, the considered contrastive loss function $\mathcal{L}_{CL}$ is formulated on image features as follows:
  \begin{align}
    \!\!\mathcal{L}_{CL}\! &=\! \left\{\!\!\!
        \begin{array}{ll}
            1 - s(\boldsymbol{f}^{i,\mathcal{T}}_1, \boldsymbol{f}^{i,\mathcal{T}}_2)),\!\!\! & \boldsymbol{y}_i^{\mathcal{T}} = 1 \\
            \max{(0, s(\boldsymbol{f}^{i,\mathcal{T}}_1, \boldsymbol{f}^{i,\mathcal{T}}_2)\!-\!m}),\!\!\! & 
            \boldsymbol{y}_i^{\mathcal{T}} = 0
        \end{array}
    \right.
    \label{EQ:eq1}
 \end{align}
where $m$ is the margin parameter and $s(\cdot, \cdot)$ measures cosine similarity.

Once the network is trained on a given training set, CBIR is performed by comparing the feature of the given query with that of each image in the archive $\mathcal{I}$. We would like to note that the proposed ANNEAL is independent from the considered metric learning loss function. Thus, instead of (1) different pairwise loss functions such as NT-Xent \cite{sohn2016improved_NTX} can be also utilized.

\subsection{ANNEAL: Uncertainty Criterion}
The first step of the proposed ANNEAL method is devoted to the selection of the most uncertain image pairs. To assess the uncertainty of image pairs, we introduce two different algorithms: 1) metric guided uncertainty estimation (MGUE); and 2) binary classifier guided uncertainty estimation (BCGUE). Each algorithm is combined with the considered diversity criterion. ANNEAL with MGUE algorithm is denoted as ANNEAL-MGUE, whereas ANNEAL with BCGUE algorithm is denoted as ANNEAL-BCGUE.

\subsubsection{Metric Guided Uncertainty Estimation (MGUE)} 
The proposed MGUE algorithm aims at assessing uncertainty of the image pairs directly in the metric space. To this end, we estimate a threshold value to distinguish between similar and dissimilar images. This is automatically achieved based on the distances between similar and dissimilar images in the feature space. If the similarity in the feature space between any two images exceeds the threshold value, then the images are considered as similar, vice versa. The image pairs that have a cosine similarity value, which is closest to the estimated threshold value, are the most uncertain ones. Let $\alpha_{t}$ be the estimated threshold value. Then, the similarity of an image pair $\boldsymbol{X}_i = (\boldsymbol{I}^{i}_1,\boldsymbol{I}^{i}_2)$ can be determined as follows:
  \begin{align}
    \!\!{X}_i = (\boldsymbol{I}^{i}_1,\boldsymbol{I}^{i}_2)\! &=\! \left\{\!\!\!
        \begin{array}{ll}
            \text{similar},\!\!\! & \text{if} \ \ s(\boldsymbol{f}^{i}_1, \boldsymbol{f}^{i}_2) \geq \alpha_{t}  \\
            \text{dissimilar},\!\!\! & \text{if} \ \
            s(\boldsymbol{f}^{i}_1, \boldsymbol{f}^{i}_2) < \alpha_{t}
        \end{array}
    \right.
    \label{EQ:similarity_threshold_concept}
 \end{align}
To estimate the threshold value $\alpha_{t}$, we calculate the mean and the standard deviation of similar and dissimilar image pairs in the current training set as follows: 
\begin{align}
    \label{eq:mean_dis}
    \mu_{sim} = \frac{1}{S} \sum_{i=1}^{S} s(\boldsymbol{f}^{i,\mathcal{T}}_{1}, \boldsymbol{f}^{i,\mathcal{T}}_2), && \boldsymbol{y}_i^{\mathcal{T}} = 1
\end{align}

\begin{align}
    \label{eq:std_similar}
    \sigma_{sim} = \sqrt{\frac{1}{S} \sum_{i=1}^{S} (s(\boldsymbol{f}^{i,\mathcal{T}}_{1}, \boldsymbol{f}^{i,\mathcal{T}}_2)-\mu_{sim})^2}, && \boldsymbol{y}_i^{\mathcal{T}} = 1
\end{align}

\begin{align}
    \label{eq:mean_dis}
    \mu_{dsim} = \frac{1}{D} \sum_{i=1}^{D} s(\boldsymbol{f}^{i,\mathcal{T}}_{1}, \boldsymbol{f}^{i,\mathcal{T}}_2), && \boldsymbol{y}_i^{\mathcal{T}} = 0
\end{align}

\begin{align}
    \label{eq:std_dis}
    \sigma_{dsim} = \sqrt{\frac{1}{D} \sum_{i=1}^{D} (s(\boldsymbol{f}^{i,\mathcal{T}}_{1}, \boldsymbol{f}^{i,\mathcal{T}}_2)-\mu_{dsim})^2},&& \boldsymbol{y}_i^{\mathcal{T}} = 0
\end{align}
where $\mu_{sim}$,$\sigma_{sim}$ and $\mu_{dsim}$,$\sigma_{dsim}$ are the mean and the standard deviation of the cosine similarity between the images composing the similar and dissimilar pairs, respectively. $S$ and $D$ are the number of total similar and dissimilar pairs in the training set, respectively. Based on these statistics, the threshold value $\alpha_{t}$ is defined as follows:
\begin{align}
    \label{eq:threshold}
    \alpha_{t} = \frac{\mu_{sim} + \mu_{dsim}-\lambda (\sigma_{sim} -\sigma_{dsim})}{2},
\end{align}
where $\lambda $ is a hyper-parameter that defines the width of the standard deviation. The closer the cosine similarity between the images that compose the unlabeled pairs are to the customized threshold $\alpha_{t}$ the more uncertain the image pairs are. We define the uncertainty of a given unlabeled pair $\boldsymbol{X}_i^\mathcal{U} = (\boldsymbol{I}^{i,\mathcal{U}}_1,\boldsymbol{I}^{i,\mathcal{U}}_2)\in \mathcal{U}$ as the absolute difference between the cosine similarity of the image pair in the feature space and the threshold value $\alpha_{t}$, as follows: 
\begin{align}
    \label{eq:uncertain}
    Unc(\boldsymbol{X}_i^{\mathcal{U}}) =  |s(\boldsymbol{f}^{i,\mathcal{U}}_{1}, \boldsymbol{f}^{i,\mathcal{U}}_2)-\alpha_{t}|.
\end{align}
We calculate the uncertainty of each pair in the unlabeled set $\mathcal{U}$, and then sort the pairs in ascending order and we select the set $\mathcal{S}^{Unc}=\{\boldsymbol{X}_1,\boldsymbol{X}_2,\ldots,\boldsymbol{X}_p\}$ of $p$ most uncertain image pairs.

\subsubsection{Binary Classifier Guided Uncertainty Estimation (BCGUE)}
The proposed BCGUE algorithm aims at assessing the uncertainty of the image pairs based on the confidence of a binary classifier used to classify image pairs as similar/dissimilar with respect to each other. The lower the confidence of the classifier in assigning the similarity label to an image pair, the higher is the uncertainty of the pair. To this end, we employ a binary classifier after the considered SNN to classify images as similar or dissimilar. The binary classifier consists of three fully connected layers, while the last layer is made up of a single neuron. We obtain the class posterior probability of $\boldsymbol{y}_i^{\mathcal{T}}$, for a given image pair $\boldsymbol{X}_i^{\mathcal{T}}$, activating the single neuron of the last layer with \textit{sigmoid} activation function as follows:
\begin{align}
    \label{eq:posterior}
    P(\boldsymbol{y}_i^{\mathcal{T}} | \boldsymbol{X}_i^{\mathcal{T}} ) = \frac{1}{1+e^{-\phi(g(\boldsymbol{X}_i^{\mathcal{T}}))}},
\end{align}
where $g$ and $\phi$ represent the SNN and the binary classifier, respectively. To optimize the model parameters, in addition to $\mathcal{L}_{CL}$, we employ the binary cross entropy loss function $\mathcal{L}_{BCE}$ as follows:
\begin{align}
    \mathcal{L}_{BCE} &= \boldsymbol{y}_i^{\mathcal{T}}\log(P(\boldsymbol{y}_i^{\mathcal{T}} | \boldsymbol{X}_i^{\mathcal{T}})) \nonumber \\
    &\quad + (1-\boldsymbol{y}_i^{\mathcal{T}})\log(1-P(\boldsymbol{y}_i^{\mathcal{T}} | \boldsymbol{X}_i^{\mathcal{T}})).
    \label{EQ:bce}
\end{align}
We train the whole system end-to-end combining $\mathcal{L}_{CL}$ and $\mathcal{L}_{BCE}$ as follows:
 \begin{align}
    \mathcal{L} =  (1-\gamma)\mathcal{L}_{CL}+\gamma\mathcal{L}_{BCE},
 \end{align} 
where $\gamma$ is the balancing factor.
 
Once the model is trained, we use the posterior class probabilities given by \mbox{(\ref{eq:posterior})} to evaluate the confidence of the image pairs. Specifically, for a binary classification problem the most uncertain image pairs are the ones characterized by the posterior class probability $P(\boldsymbol{y}_i^{\mathcal{U}} | \boldsymbol{X}_i^{\mathcal{U}} )\approx 0.5 $. We calculate the uncertainty of each pair in the unlabeled set $\mathcal{U}$ using \mbox{(\ref{eq:uncertain})}, where we substitute the cosine similarity between image pairs $s(\boldsymbol{f}^{i,\mathcal{U}}_{1}, \boldsymbol{f}^{i,\mathcal{U}}_2)$ with the class posterior probabilities $P(\boldsymbol{y}_i^{\mathcal{U}} | \boldsymbol{X}_i^{\mathcal{U}})$ and set the value of $\alpha_{t}=0.5$. Then, we sort the image pairs in ascending order and we select the set $\mathcal{S}^{Unc}=\{\boldsymbol{X}_1,\boldsymbol{X}_2,\ldots,\boldsymbol{X}_p\}$ of $p$ most uncertain image pairs.

\subsection{ANNEAL: Diversity Criterion}
The second step of the proposed ANNEAL method is devoted to the selection of the most diverse image pairs among the uncertain ones. To this end, we exploit a clustering-based strategy to first partition the uncertain pairs into $h$ sets (i.e., clusters). We select the most uncertain pair from each cluster to obtain a set $\mathcal{S}=\{\boldsymbol{X}_1,\boldsymbol{X}_2,\ldots,\boldsymbol{X}_h\}$ of $h$ image pairs that are at the same time the most uncertain and diverse ones. As a clustering strategy we use k-means clustering algorithm presented in \cite{jain1988algorithms_clustering}. However, any clustering approach can be used in the proposed method.

After the selection of the most informative image pairs, a human expert annotates each pair as similar or dissimilar and the labeled pairs are added to the training set. We further enrich the training set with zero annotation cost exploiting the transitivity property of similarity \mbox{\cite{roy2018exploiting_transitive}}.
Let $\boldsymbol{X}_i^{\mathcal{T}}$ and $\boldsymbol{X}_j^{\mathcal{T}}$ be two image pairs that share a common image (i.e., same image is present in both pairs). Let $\boldsymbol{I}^{i,\mathcal{T}}_2 = \boldsymbol{I}^{j,\mathcal{T}}_2$ be the shared image between the pairs $\boldsymbol{X}_i^{\mathcal{T}}$ and $\boldsymbol{X}_j^{\mathcal{T}}$. If $\boldsymbol{X}_i^{\mathcal{T}}$ and $\boldsymbol{X}_j^{\mathcal{T}}$ are both similar pairs (i.e.,  $\boldsymbol{y}_i^{\mathcal{T}} =1$ and $\boldsymbol{y}_j^{\mathcal{T}} = 1$) then from transitivity property we form a new similar pair $\boldsymbol{X}_k^{\mathcal{T}} = (\boldsymbol{I}^{i,\mathcal{T}}_1,\boldsymbol{I}^{j,\mathcal{T}}_1)$ with $\boldsymbol{y}_k^{\mathcal{T}} = 1$. Similarly, if  $\boldsymbol{X}_i^{\mathcal{T}}$ is a similar pair and $\boldsymbol{X}_j^{\mathcal{T}}$ is a dissimilar pair (i.e.,  $\boldsymbol{y}_i^{\mathcal{T}} =1$ and $\boldsymbol{y}_j^{\mathcal{T}} = 0$) then from transitivity property we can infer that $\boldsymbol{X}_k^{\mathcal{T}} = (\boldsymbol{I}^{i,\mathcal{T}}_1,\boldsymbol{I}^{j,\mathcal{T}}_1)$ is a dissimilar pair and can be labeled with $\boldsymbol{y}_k^{\mathcal{T}} = 0$. If instead $\boldsymbol{X}_i^{\mathcal{T}}$ and $\boldsymbol{X}_j^{\mathcal{T}}$ are both dissimilar pairs (i.e.,  $\boldsymbol{y}_i^{\mathcal{T}} =0$ and $\boldsymbol{y}_j^{\mathcal{T}} = 0$) we cannot infer any information about the pair $\boldsymbol{X}_k^{\mathcal{T}} = (\boldsymbol{I}^{i,\mathcal{T}}_1,\boldsymbol{I}^{j,\mathcal{T}}_1)$. As a result, no new pair will be formed in this case. It is worth noting that we only perform a single transitive step  and we do not proceed another transitive step with the newly generated pairs.
\section{Dataset Description and Experimental Design} \label{dataset_setup}

\subsection{Datasets}
To asses the performance of the proposed cost-efficient AL method, we carried out different experiments on two different archives that consists of VHR images. 

The first archive, namely UC-Merced, consists of 2100 images selected from aerial ortho-imagery \cite{yang2010bag_UCM} and has in total 21 different categories: agricultural, airplane, baseball diamond, beach, buildings, chaparral, dense residential, forest, freeway, golf course, harbor, intersection, medium density residential, mobile home park, overpass, parking lot, river, runway, sparse residential, storage tanks, and tennis courts. The images are of size $256$x$256$ pixels with a spatial resolution of 30 cm and are downloaded from United States Geological Survey (USGS) National Map of the following U.S. regions: Birmingham, Boston, Buffalo, Columbus, Dallas, Harrisburg, Houston, Jacksonville, Las Vegas, Los Angeles, Miami, Napa, New York, Reno, San Diego, Santa Barbara, Seattle, Tampa, Tucson, and Ventura.

The second archive for which we carried out our experiments is the Aerial Image Dataset (AID) \cite{xia2017aid} which consists of 10000 images collected from Google Earth imagery and has in total 30 different categories: airport, bare land, baseball field, beach, bridge, center, church, commercial, dense residential, desert, farmland, forest, industrial, meadow, medium residential, mountain, park, parking, playground, pond, port, railway station, resort, river, school, sparse residential, square, stadium, storage tanks and viaduct. \cite{xia2017aid}. The number of images per category varies from 220 up to 420 images that are chosen from different regions around the world, mainly from the United States, England, France, Italy, Japan, Germany, etc, under different times, seasons and imaging conditions . The images are of size $600$x$600$ pixels and have different spatial resolutions in the range between 50 to 80 cm.

\subsection{Experimental Setup}
In all our experiments we have randomly divided the two archives into three sets: training (80\%), validation (10\%) and test (10\%). To derive the initial training set $\mathcal{T}$ we randomly select $5\%$ and $1\%$ of the images for UC-Merced and AID datasets, respectively and then create the pairs by randomly choosing four similar and four dissimilar images for each labeled images based on their categories (i.e., class labels). The unlabeled set $\mathcal{U}$ consists of all training pairs that are not part of $\mathcal{T}$. At each AL iteration we extend the initial training set $\mathcal{T}$ by adding 336 and 392 most informative image pairs from $\mathcal{U}$ of UC-Merced and AID, respectively. This means that 336 and 392 bits of information are included at each AL iteration of UC-Merced and AID, respectively. This corresponds to the annotation of 5\% and 1\% of the images with LULC class labels for UC-Merced and AID archives, respectively, considering the datasets sizes (AID dataset is almost five times bigger than UC-Merced). ResNet18 \cite{he2016deepResnet} is chosen as a backbone of the image characterization module (i.e., SNN) with initial weights obtained from the pre-trained model on ImageNet \cite{ImageNett}. Subsequently the model's output is projected through a non-linear projection head composed of two fully connected layers with hidden dimension of 512 and 256 units and RelU activation for both the algorithms. The weights of the projection head are randomly initialized. In the case of the BCGUE algorithm, the output of the ResNet18 is concatenated and fed into the binary classifier. The model is trained for 15 epochs using a batch size of 128 and Adam Optimizer \cite{kingma2014adam} with a learning rate of $10^{-4}$. For the retrieval, the projection head is discarded and the feature representations obtained from the backbone (i.e., ResNet18) are used to perform CBIR. The retrieval is performed on the test set, while the query images are taken from the validation set. The retrieval performance is measured using the widely used mean Average Precision (mAP) metric denoted as follows:
\begin{align}
    \label{eq:maP}
    mAP = \frac{1}{Q}\sum_{i=1}^{Q}\frac{1}{r_i}\sum_{j=1}^{k}Prec_i(j)\cdot \delta_i(j),
\end{align}
where $Q$ is the size of the query set; $r_i$ is the number of items that are related to the $i^{th}$ query image; $k$ is the number of images in the retrieval set; $Prec_i(j)$ is the precision at $j^{th}$ position for the $i^{th}$ query image and $\delta_i(j)$ is a binary relevance function that returns 1 if the item at $j^{th}$ position is relevant to the $i^{th}$ query image and 0 otherwise. In our experiments, the retrieval performance is measured using mAP on 5 retrieved images (mAP@5).

All the experimental results refer to the average retrieval performances obtained in three trials, while an initial training set was randomly reconstructed in each trial. As suggested in \cite{demir2014novel_AL_CBIR} we set $p = 4h$, whereas the margin $m$ parameter of the loss function is chosen to be 0.5 as default. While constructing the image pairs, there are more dissimilar pairs compared to similar ones. Approximately, $5\%$ of the pairs are similar and $95\%$ are dissimilar. To stabilize the training, we apply oversampling strategy on the minority class presented in \cite{branco2016survey}. 

We train the CBIR system on the entire training set to have an upper bound on the retrieval performance. We compare the proposed method with the following: 1) randomly selecting the image pairs at each AL iteration (i.e. random selection) and 2) classification based AL (CAL) \mbox{\cite{class_AL_2014}} for CBIR, where uncertain images are selected to be labeled with LULC class labels. In CAL, the uncertainty of an image is assessed based on the confidence of the supervised classifier in correctly classifying it with one of the LULC class labels. To make the methods comparable we use the same initial training set, the same backbone, projection head and incorporate the same number of bits of information to the labeled set at each AL iteration. For CAL, a fully-connected layer is added on top of the projection head with softmax activation, where the number of neurons is equal to the number of classes and cross-entropy loss function is used to train the model. Since CAL only evaluates uncertainty, we added a diversity criterion based on a clustering strategy to allow for a fairer comparison.

\section{Experimental Results} \label{experimental_results}
We carried out different kinds of experiments to: 1) perform a sensitivity analysis of the proposed MGUE algorithm with respect to the selection of the hyparameter $\lambda $; 2) analyse the effect of the diversity criterion on the proposed ANNEAL method; and 3) compare the proposed ANNEAL method with CAL method and random selection.
\subsection{Analysis of the Effect of the Hyparameter $\lambda $ on the Proposed MGUE Algorithm}
To assess the uncertainty of the image pairs through the proposed MGUE algorithm the threshold value needs to be estimated. The threshold value is used by the proposed MGUE algorithm to select the most uncertain image pairs, specifically those that have a cosine similarity value closest to the estimated threshold value. The threshold value is estimated based on the mean and the standard deviation of the cosine similarity values between the images that compose the similar and dissimilar pairs and contains one hyperparameter, $\lambda $ (see  (\ref{eq:threshold}))]. This hyperparameter defines the width of the standard deviation (dispersion of the cosine similarity values between the images of similar and dissimilar pairs) and should be carefully selected.

\begin{figure}
    \centering
    \includegraphics[scale =0.5]{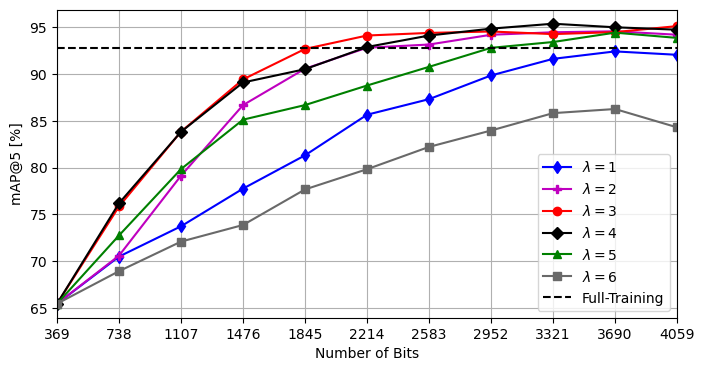}
    \caption{The average retrieval performance in terms of mAP@5 versus the number of bits of information obtained by the proposed MGUE for different values of $\lambda$ for the UC-Merced dataset.}
    \label{Fig:eta_UCM}
\end{figure}

\begin{figure}
    \centering
    \includegraphics[scale =0.5]{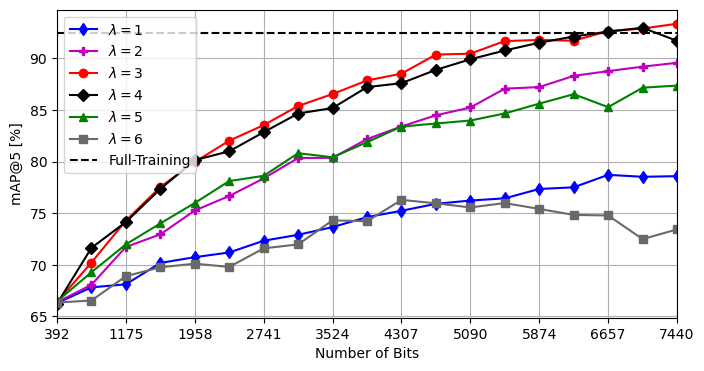}
    \caption{The average retrieval performance in terms of mAP@5 versus the number of bits of information obtained by the proposed MGUE for different values of $\lambda$ for the AID dataset.}
    \label{Fig:eta_AID}
\end{figure}

We evaluated the effect of the hyperparameter $\lambda$ used in proposed MGUE algorithm, while it is varied as $\lambda \in \{1,2,3,4,5,6\}$. Fig. ~\ref{Fig:eta_UCM} and ~\ref{Fig:eta_AID} show the retrieval performance in terms of mAP@5 versus the bits of information used and the hyperparameter $\lambda $ variation when the proposed MGUE algorithm is applied on UC-Merced and AID datasets, respectively. From the figures, one can observe that for both datasets when $\lambda =3$ the proposed MGUE algorithm achieves the best trade-off between the number of bits used for the training and the retrieval performance. When $\lambda $ increases ($\lambda >3$) or decreases ($\lambda <3$) the retrieval performance starts to drop. 
This is due to the fact that very high threshold values result in selecting easy and non informative dissimilar pairs whereas very low threshold values result in selecting easy and non informative similar pairs. Based on these results for all the subsequent experiments we fix the hyperparameter $\lambda =3$ for the ANNEAL-MGUE.

\subsection{Analysis of the Effect of Diversity Criterion}
\begin{figure}
    \centering
    \includegraphics[scale =0.5]{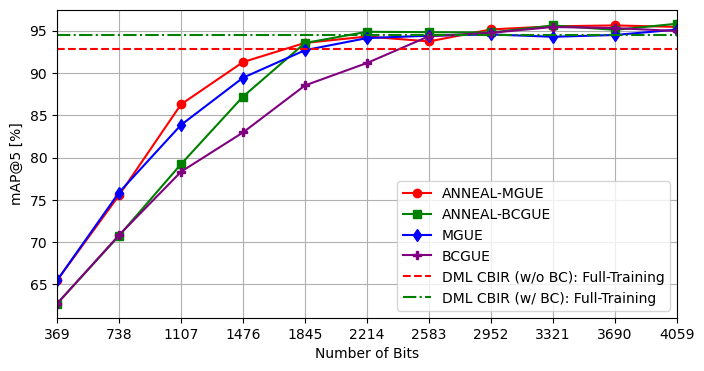}
    \caption{The average retrieval performance in terms of mAP@5 versus the number of bits of information obtained by the ANNEAL-MGUE, the ANNEAL-BCGUE and also direct use of the MGUE and the BCGUE without diversity criterion for the UC-Merced dataset.}

    \label{Fig:UD_UCM}
\end{figure}

\begin{figure}
    \centering
    \includegraphics[scale =0.5]{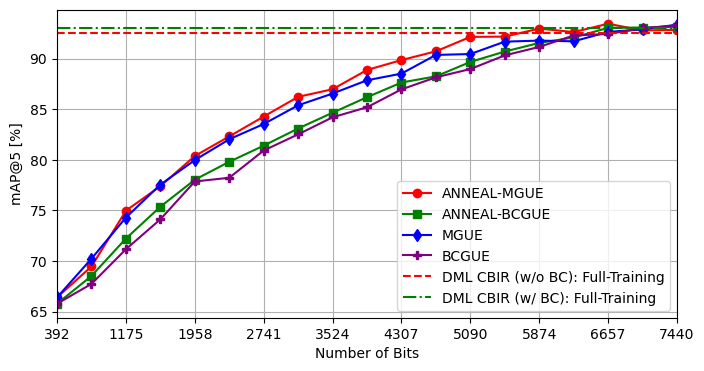}
    \caption{The average retrieval performance in terms of mAP@5 versus the number of bits of information obtained by the ANNEAL-MGUE, the ANNEAL-BCGUE and also direct use of the MGUE and the BCGUE without diversity criterion for the the AID dataset.}

    \label{Fig:UD_AID}
\end{figure}

In this subsection, we analyse the retrieval performance obtained by using only the uncertainty criterion and the combination of uncertainty criterion with diversity criterion for ANNEAL. Fig.~\ref{Fig:UD_UCM} and ~\ref{Fig:UD_AID} show the average retrieval performance in terms of mAP@5 versus the number of bits of information obtained by the ANNEAL-MGUE, the ANNEAL-BCGUE and also direct use of  MGUE and BCGUE without diversity criterion for the UC-Merced dataset and AID dataset, respectively. One can observe that for both datasets the combination of uncertainty and diversity criteria achieves the best results under different bits of information used for training. In particular, this  combination is much important and effective for ANNEAL-BCGUE. As it can be seen in Fig.~\ref{Fig:UD_UCM}, the proposed ANNEAL-BCGUE, not only converges faster to the upper bound compared to BCGUE algorithm but also shows higher retrieval results under different bits of information. This is particularly visible where the number of bits of information used for training ranges between 1107 to 2583, when the convergence to the upper bound is reached. The same behavior with a smaller difference can also be seen with ANNEAL-MGUE. 

\subsection{Comparison among the Proposed ANNEAL and Literature Methods}
\begin{figure}
    \centering
    \includegraphics[scale =0.5]{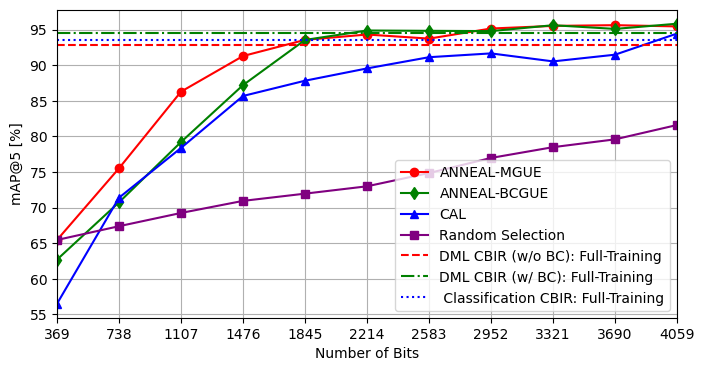}
    \caption{Comparison results between the proposed ANNEAL, CAL and random selection methods on UC-Merced dataset.}
    \label{Fig:comp_UCM}
\end{figure}

\begin{figure}
    \centering
    \includegraphics[scale =0.5]{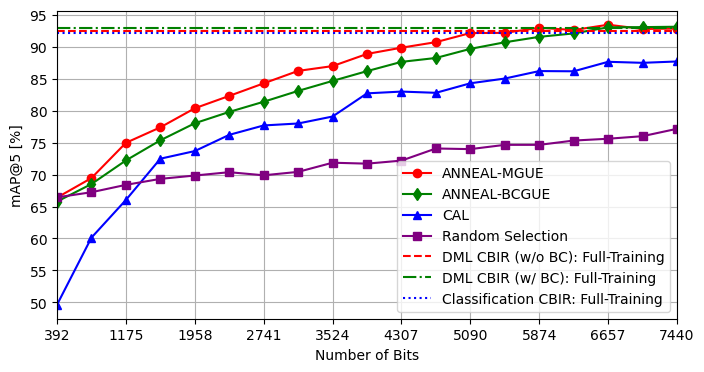}
    \caption{Comparison results between the proposed ANNEAL, CAL and random selection methods on AID dataset.}
    \label{Fig:comp_AID}
\end{figure}

In this subsection, we compare the proposed ANNEAL method with CAL \mbox{\cite{class_AL_2014}} and random selection. Fig. \mbox{~\ref{Fig:comp_UCM}} and Fig. \mbox{~\ref{Fig:comp_AID}} depict the retrieval performance versus the number of bits of information used for training obtained by the proposed ANNEAL-MGUE and ANNEAL-BCGUE methods, CAL \mbox{\cite{class_AL_2014}} and random selection for UC-Merced and AID datasets, respectively. By analyzing the figures, one can see that ANNEAL-MGUE and ANNEAL-BCGUE lead to the highest retrieval performance for all the iterations and outperform the CAL and random selection methods for both datasets. As an example, for the UC-Merced dataset (see Fig. \mbox{~\ref{Fig:comp_UCM}}), when 1845 bits of information are used for training the ANNEAL-MGUE and ANNEAL-BCGUE provide an average retrieval performance of 93.56\%, 93.55\% in terms of mAP@5, respectively, whereas those obtained by CAL and random selection under the same bits of information are of 87.82\% and 71.96\%, respectively. Moreover, to reach to the same retrieval performance of ANNEAL-MGUE and ANNEAL-BCGUE, the CAL method requires 4059 bits of information for training (i.e., final training size). This corresponds to 6 more AL iterations compared to the proposed ANNEAL-MGUE and ANNEAL-BCGUE methods. Another important result is that ANNEAL-MGUE and ANNEAL-BCGUE reach convergence two times faster than the CAL method. In particular, the convergence for both ANNEAL-MGUE and ANNEAL-BCGUE is reached when using only 2214 bits of information.
For random selection, the retrieval performance obtained with the final training set does not reach convergence. 

By analyzing Fig. \mbox{~\ref{Fig:comp_AID}} one can observe that the retrieval performances of the ANNEAL-MGUE and ANNEAL-BCGUE methods are in general better than those of CAL and significantly better than random selection under different bits of information used for training. For example, the retrieval performances for ANNEAL-MGUE and ANNEAL-BCGUE obtained with the final size of the training set (7440 bits) is of 92.85\% and of 93.13\% respectively, whereas the ones obtained by CAL method and random selection are of 87.70\% and 77.16\% respectively. One can observe that ANNEAL-MGUE can reach the final training size performance of the CAL method using less than half bits (i.e., 3524) of information for training whereas ANNEAL-BCGUE requires 4307 (slightly more than half) bits of information. Another important observation is that, while ANNEAL-MGUE and ANNEAL-BCGUE reach the convergence using 5482 and 6266 bits of information for training, respectively, the retrieval performance obtained using the final training set by the CAL and random selection methods does not reach convergence. All these results prove that ANNEAL-MGUE and ANNEAL-BCGUE are more effective than the CAL and random selection methods in creating informative training set with a lower cost.

By analyzing Fig. \mbox{~\ref{Fig:comp_UCM}} and Fig. \mbox{~\ref{Fig:comp_AID}}, one can observe that for both datasets ANNEAL-MGUE shows generally higher retrieval performances compared to ANNEAL-BCGUE under each number of bits of information used for training. This is particularly noticeable during the first AL iterations for UC-Merced dataset (see Fig. \mbox{~\ref{Fig:comp_UCM}}), where ANNEAL-MGUE retrieval performances are generally much better than ANNEAL-BCGUE. As an example, the retrieval results obtained by ANNEAL-MGUE where the number of bits of information ranges from 738 to 1476 for UC-Merced dataset (see Fig. \mbox{~\ref{Fig:comp_UCM}}) are on average 5\% higher compared to those obtained by ANNEAL-BCGUE. One can observe that this trend is confirmed also in the AID dataset for which the results obtained by ANNEAL-MGUE are on average 2\% higher compared to those obtained by ANNEAL-BCGUE. This is due to the fact that ANNEAL-MGUE is more effective in selecting uncertain pairs compared to ANNEAL-BCGUE due to the MGUE algorithm that directly operates in the metric space.

\begin{figure*}
    \centering
    \includegraphics[scale =0.6]{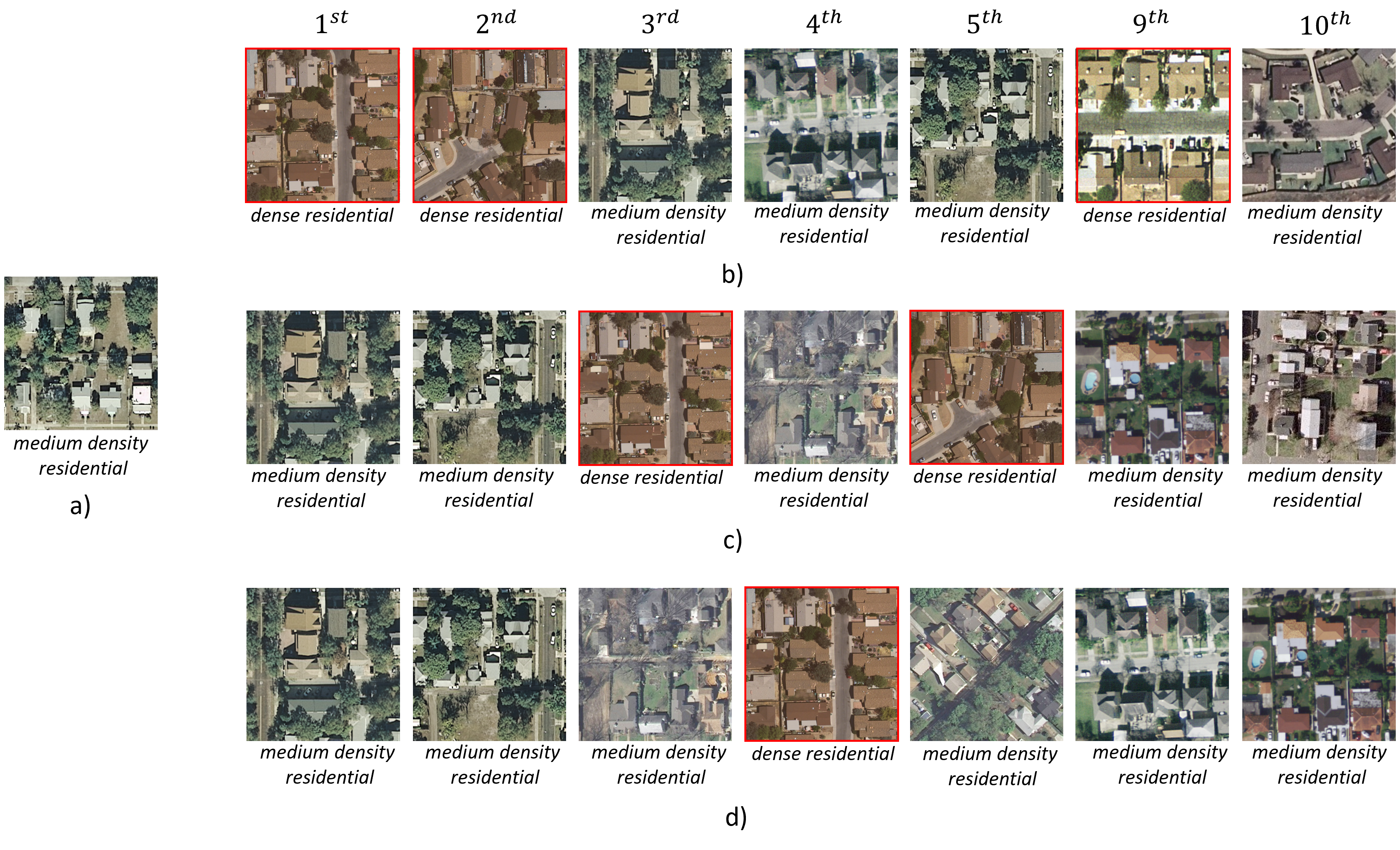}
    \caption{medium density residential image retrieval example. (a) Query image and images retrieved when 1476 bits of information are used for training the CBIR system when  b) CAL; c) the proposed ANNEAL-BCGUE; and d) the proposed ANNEAL-MGUE are applied on UC-Merced dataset. Red frames represent images that are dissimilar to the query image based on the class label.}
    \label{Fig:Medium_q_ucm}
\end{figure*}

Fig.~\ref{Fig:Medium_q_ucm} shows an example of images retrieved by the CAL method and the proposed ANNEAL-MGUE and ANNEAL-BCGUE, where the query image is selected from medium density residential category of the UC-Merced dataset when 1467 bits of information are used for training. The images that are different from the category of the query image are represented within red frames. As it can be seen in Fig.~\ref{Fig:Medium_q_ucm} (a), the CAL method retrieves the highest number of images that are different from the query category. In particular, we can see that the first two and the ninth retrieved images are from the dense residential category of the UC-Merced dataset. It is worth mentioning that the two categories, medium density residential area and dense residential area, are very similar to each other. We can see that among the images retrieved by ANNEAL-BCGUE [see Fig. ~\ref{Fig:Medium_q_ucm} (c)] two images are from the dense residential area. However, different from CAL method, those images are positioned later in the retrieval order (in the third and fourth position). In the case of ANNEAL-MGUE method [see Fig. ~\ref{Fig:Medium_q_ucm} (d)] almost all the retrieved images correspond to the category of the medium density residential. The only image that does not share the same category with the query image is the fourth retrieved one that belongs to the dense residential area.

\begin{figure*}
    \centering
    \includegraphics[scale =0.6]{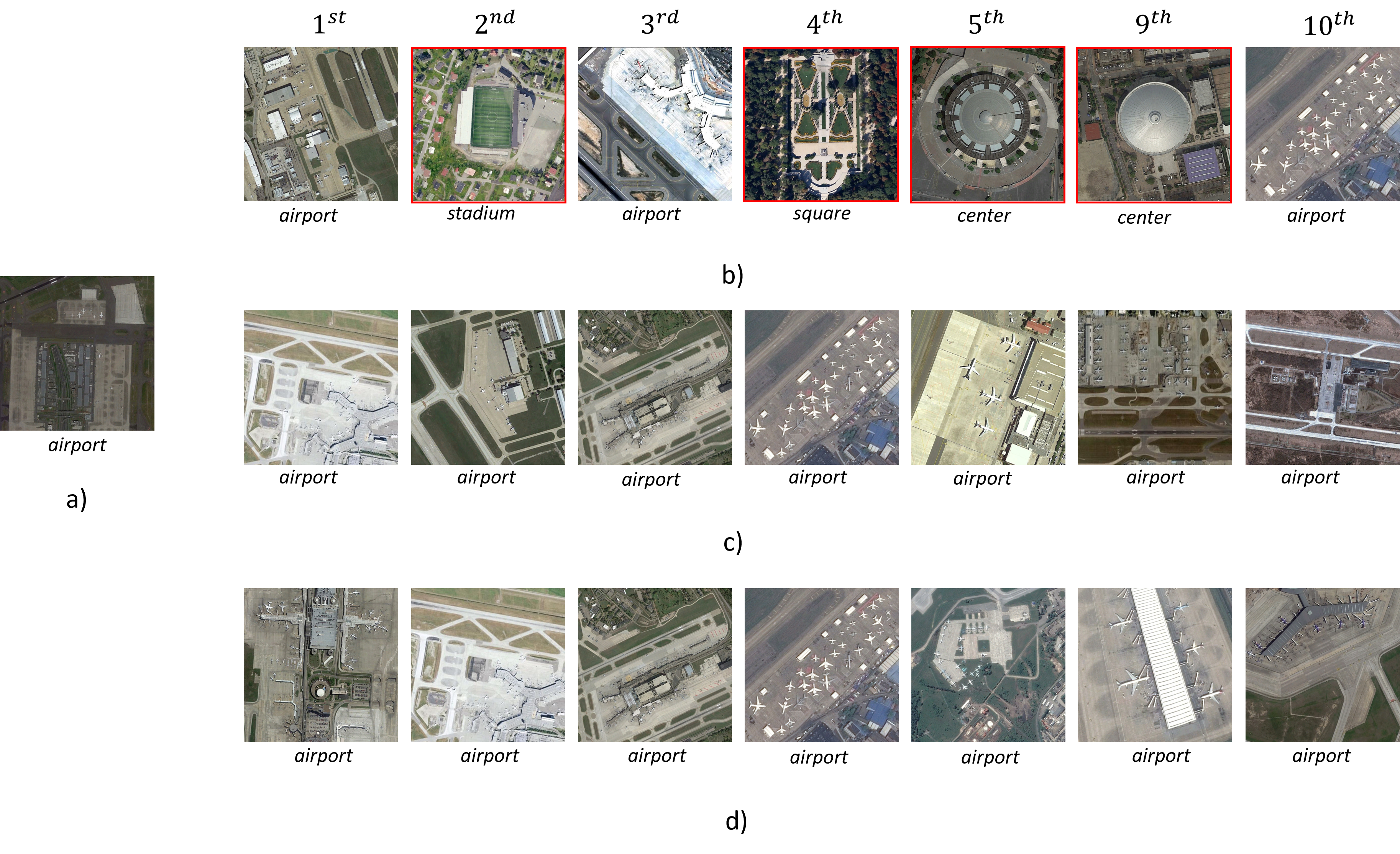}
    \caption{Airport image retrieval example. (a) Query image and images retrieved when 5090 bits of information are used for training the CBIR system when b) CAL; c) the proposed ANNEAL-BCGUE; and d) the proposed ANNEAL-MGUE are applied on the AID dataset. Red frames represent images that are dissimilar to the query image based on the class label.}
    \label{Fig:airport_q_AID}
\end{figure*}

Fig.~\ref{Fig:airport_q_AID} shows an example of images retrieved by CAL method and the proposed ANNEAL-MGUE and ANNEAL-BCGUE methods, where the query image is selected from airport category of the AID dataset when 5090 bits of information are used for training. One can observe that CAL method retrieves four images that belong to different categories with respect to the query one [see Fig. ~\ref{Fig:airport_q_AID} (b)]. We can see that both ANNEAL-MGUE and ANNEAL-BCGUE are capable of retrieving the images that share the same category as the query one. This is also reflected in the quantitative results of the AID dataset (see Fig. ~\ref{Fig:comp_AID}) where the proposed ANNEAL method achieve better performances with respect to CAL.

\section{Conclusion}\label{conclusion}
In this paper, we have presented ANNEAL that is an annotation cost-efficient AL method specifically designed for DML driven CBIR systems. The aim of the proposed AL method is to construct a small but informative training set composed of similar and dissimilar image pairs while maintaining high retrieval performance. To this end, the proposed ANNEAL method selects the most informative image pairs combining the uncertainty and diversity criteria in two consecutive steps. In the first step, to select the most uncertain image pairs we introduce two algorithms: 1) metric guided uncertainty estimation (MCGUE); and 2) binary classifier guided uncertainty estimation (BCGUE). MCGUE algorithm assesses the uncertainty of the image pairs directly in the metric space by estimating a threshold value to distinguish between similar and dissimilar image pairs based on their cosine similarity in the metric space. The closer the cosine similarity between any two images (i.e., image pairs) the higher their uncertainty. BCGUE algorithm assesses the uncertainty of the image pairs based on the confidence of a binary classifier in predicting the correct similarity label. The lower the confidence of the classifier, the higher the uncertainty of the image pair. In the second step, the proposed ANNEAL method selects the most diverse image pairs among the most uncertain ones based on k-means clustering algorithm. ANNEAL combines one of the proposed uncertainty methods with clustering strategy to select the most informative samples. ANNEAL with MGUE algorithm is denoted as ANNEAL-MGUE, while that with BCGUE algorithm as ANNEAL-BCGUE. The selected most uncertain and diverse image pairs are sent to a human expert to be annotated as being similar or dissimilar. 

Differently from previous AL methods in CBIR problems, ANNEAL is independent from the selected query image. As a result, it does not require training an ad-hoc binary classifier for each query image. This makes it more suitable for operational scenarios compared to previous AL methods designed for CBIR problems. Compared to a literature work (e.g., classification based active learning \mbox{\cite{class_AL_2014}}), for which an ad-hoc classifier is also not necessary, experimental results show the effectiveness of ANNEAL in creating a small but informative training set with a lower annotation cost, while achieving a higher CBIR accuracy. This is due to the capability of the proposed uncertainty estimation algorithms MGUE and BCGUE to assess the uncertainty of image pairs rather than single images. This allows to: i) model image similarities in a metric space (which is of great importance for CBIR problems); and ii) reduce the annotation cost of a training sample from $log_2{C}$ for $C$ classes to 1 ($log_2{2}$) in terms of bits of information. In detail, among MGUE and BCGUE algorithms experimental results show that MCGUE algorithm achieves higher performance than BCGUE algorithm. This is due to the fact that MCGUE algorithm employs uncertainty estimation directly in a deep metric space that further enhances the selection of informative image pairs, and thus increases the CBIR performance compared to BCGUE algorithm. 
We would like to point out that the way of annotating training samples with binary labels in ANNEAL can significantly decrease the annotation cost also for other RS image analysis problems that require a training set composed of image pairs. As an example, ANNEAL can be integrated for change detection problems when the label of a training sample becomes change/no-change instead of similar/dissimilar. As a future work, we plan to extend the proposed method to different RS image analysis tasks based on image pairs. Moreover, using ternary labels while annotating training samples (i.e., image triplets) can further improve the effectiveness of ANNEAL to employ DML for CBIR problems at the cost of a slight increase in annotation cost ($log_2{2}\to log_2{3}$). Accordingly, as a future development of this work, we plan to extend the proposed method to DML based CBIR systems trained with image triplets.

\section{Acknowledgement}
This work is supported by the European Research Council (ERC) through the ERC-2017-STG BigEarth Project under Grant 759764 and by the European Space Agency through the DA4DTE (Demonstrator precursor Digital Assistant interface for Digital Twin Earth) project. Julia Henkel’s scholarship is funded by Google Research.

\ifCLASSOPTIONcaptionsoff
  \newpage
\fi

\bibliography{references}
\bibliographystyle{ieeetr}
\vfill

\end{document}